\documentclass{article}
\usepackage{obs_study_style}
\usepackage[utf8]{inputenc}
\usepackage{subcaption}
\usepackage{todonotes}
\usepackage{xcolor}
\definecolor{midnightblue}{rgb}{0.1, 0.1, 0.44}
\usepackage{hyperref}
\hypersetup{
     colorlinks   = true,
     linkcolor    = black,
     citecolor    = midnightblue
}

\title{Breiman's two cultures: You don't have to choose sides}
\author{\name Andrew C.~Miller \email acmiller@apple.com \\
       \addr Apple 
       \AND
       \name Nicholas J.~Foti \email nicholas\_foti@apple.com \\
       \addr Apple
       \AND
       \name Emily B.~Fox  \email emily\_fox@apple.com \\
       \addr Apple}
\date{}
\begin{document}

\maketitle

\begin{abstract}
Breiman's classic paper casts data analysis as a choice between two cultures: \emph{data modelers} and \emph{algorithmic modelers}. Stated broadly, data modelers use simple, interpretable models with well-understood theoretical properties to analyze data. Algorithmic modelers prioritize predictive accuracy and use more flexible function approximations to analyze data.  This dichotomy overlooks a third set of models --- mechanistic models derived from scientific theories (e.g., ODE/SDE simulators). Mechanistic models encode application-specific scientific knowledge about the data. And while these categories represent extreme points in model space, modern computational and algorithmic tools enable us to interpolate between these points, producing flexible, interpretable, and scientifically-informed hybrids that can enjoy accurate and robust predictions, and resolve issues with data analysis that Breiman describes, such as the Rashomon effect and Occam's dilemma. Challenges still remain in finding an appropriate point in model space, with many choices on how to compose model components and the degree to which each component informs inferences.
\end{abstract}


\paragraph{Introduction.}
In numerous areas, and especially in scientific disciplines, we face the challenge of drawing inferences from limited observations of complex phenomena, complicating the application of standard machine learning tools. For example, health applications present a number of issues that frustrate the formation of robust predictions and inferences. Observations are often limited to a small subset of variables at play in an intricate physiological process. Clinical studies face challenges in recruitment, compliance, and cost, often resulting in small samples. Observational studies can feature heavy confounding, inducing correlations that may disappear in new contexts. Biomarkers can be governed by unknown physiological mechanisms that drive their short- and long-term dynamics. Given such messy data, we are then tasked with the challenge of specifying a model. On the one hand, we want flexible models to capture noisy, complex processes.  On the other hand, we need to incorporate an inductive bias to handle limited data or weakly identified scenarios; ideally, this inductive bias leverages our scientific understanding and ensures that predictions are physiologically plausible and clinically sensible. Through this lens of analyzing health data we revisit Breiman's classic ``Two Cultures'' paper.

Breiman describes data analysis as a choice between two cultures: \emph{data modelers} and \emph{algorithmic modelers}.
Broadly, data modelers use simple and interpretable models with well-understood theoretical properties --- e.g., linear Gaussian models.
Algorithmic modelers, on the other hand, prioritize predictive accuracy and use more flexible models to find patterns and make predictions --- e.g., random forests and neural networks.
Despite benefits and shortcomings of both frameworks, Breiman makes a strong case for using algorithmic models to tackle real-world data problems.
Algorithmic models are appealing for their flexibility and potential to capture the complex dynamics and relationships at play; however, these models can suffer from under-identifiability, even in the presence of a lot of data, leading to implausible predictions and inferences.  
Data models, in contrast, can be simpler to fit to data --- both computationally and in terms of sample complexity --- and provide the allure of intepretability.  As Breiman discusses, however, such beneficial properties have limited value if the model is fundamentally ill-suited to describe the data at hand. 

Breiman's dichotomy overlooks an alternative approach to modeling data. 
In some cases, the underlying data generating process is well-understood and approximately described by a \emph{mechanistic model}. Mechanistic models, derived from scientific theories, encode application-specific knowledge about the causal mechanisms that generate the observed data --- e.g., through a set of ordinary or partial differential equations.  
Where data models embrace simplicity and algorithmic models flexibility, mechanistic models confer an inductive bias informed by domain-specific scientific knowledge.
Breiman takes aim at the statistics and machine learning communities, but the inclusion of mechanistic models folds in another community of modelers and data analyzers, namely scientists developing and testing theories.

These three modeling frameworks bring distinct benefits and drawbacks, naturally raising the question, can we enjoy the benefits of each instead of simply choosing one? 
Growing interest has been in researching and developing such \emph{hybrid models} --- models that lie on the continuum between \emph{data}, \emph{algorithmic}, and \emph{mechanistic} models. Due to the tremendous computational and algorithmic advances made in the past 20 years since Breiman's paper, we are no longer faced with choosing between a few discrete modeling frameworks. Automatic differentiation tools \citep{tensorflow2015-whitepaper, jax2018github, pytorch, team2016theano}, approximate inference methods \citep{blei2017variational, hoffman2014no, kucukelbir2016automatic, ranganath2014black}, probabilistic programming languages \citep{bingham2019pyro, carpenter2017stan, salvatier2016probabilistic, tran2016edward}, and advances in causal inference \citep{bollen2014structural, hernan2020causal, pearl2009causality} together enable us to specify models drawing from a full menu of tools, including flexible (i.e., algorithmic model), simple (i.e., data model), and highly structured (i.e., mechanistic model) components. This more comprehensive space of model choices grants finer control over issues of interpretability and nuisance variation. By interpolating between the extremes of the cultures, we can more precisely tailor a solution to the particular scientific pursuit at hand.

Hybrid models raise new challenges, however.  What variation in our data can be informed by scientific understanding vs.~data-driven structure?  How do we specify the intersection of components?  How can we efficiently perform statistical inference and make predictions? 
In this note, we examine hybrid models in the statistics and machine learning literature framed within our ontology, highlight the issues that motivate their development, and describe the tools that enable their implementation.

\paragraph{The extreme points of model space.}
By Breiman's definition, data models tend to be simple, easy to interpret, and theoretically tractable, but have limited expressivity. 
This framework describes a wide set of standard statistical tools, including linear regression, logistic regression, autoregressive moving average (ARMA) models \citep{box2015time}, generalized linear models \citep{nelder1972generalized}, and linear mixed models \citep{mcculloch2005generalized}. Due to the previously mentioned properties, data models are the go-to methods in many fields, such as econometrics, political science, and other social sciences.

Algorithmic models, by contrast, are highly flexible function approximators that learn predictive features (or representations) from data. Tools in this class include neural networks \citep{rumelhart1985learning}, random forests \citep{breiman2001random}, and more complex deep learning models, long short-term memory networks (LSTMs) \citep{hochreiter1997long}, and transformers \citep{vaswani2017attention}. Algorithmic models have undergone a large shift in the last 20 years with deep neural networks becoming a dominant modeling approach due to hardware advances and the availability of large datasets.

Mechanistic models are built from first principles, encoding scientific knowledge about the underlying phenomena that produces the observed data.
For example, differential equation models in epidemiology --- e.g., the \emph{susceptible-exposed-infectious-removed} (SEIR) framework --- describe the dynamics of an infectious disease spreading throughout a population \citep{anderson1992infectious, kermack1927contribution}. 
Other examples from the biomedical literature include models of insulin-glucose dynamics in diabetics \citep{bergman1981physiologic, dalla2002oral, dalla2007meal, man2014uva} and cardiovascular function \citep{ kotani2005model, mcsharry2003dynamical, trayanova2011whole}. 
While these models are interpretable to domain experts and encode domain knowledge in fine detail, they can be difficult to fit to data or extend to new types of observations. Additionally, mechanistic models have their own assumptions that must be checked.

These three model classes are not exhaustive.  Instead, we consider them a set of building blocks that can be combined in various ways to address a variety of scientific questions and prediction problems.  Each with distinct strengths and weaknesses, one model class may not suitably address the complexities of the scientific problem at hand.

Overly simple data models, as Breiman argues, may not have the flexibility to capture all of the subtle variation present in the data required to make good predictions. 
Algorithmic models can be uninterpretable, making it difficult to explain, audit or critique the predictions from a complex neural network or random forest. 
Overly flexible models are data-hungry, potentially requiring many more observations than a simpler model to obtain comparable predictions.

One major shortcoming of both data and algorithmic models is that they are, fundamentally, driven by correlations. 
Predictions are rooted in statistical correlations between inputs and outputs present in the training data.
This causal agnosticism, coupled with over-parameterized algorithmic models, leads to what Breiman calls the Rashomon effect: There are many parameter settings that lead to comparable predictive performance.
And despite the profound evolution of neural network architectures and learning algorithms, this kind of under-specification is common, even in machine learning systems with access to the largest datasets available \citep{d2020underspecification}.
Compounding this issue, we may have access to observational data in one context, but must make predictions in a different, unseen context.  In health applications, a common goal is to predict the occurrence of rare events, which, by definition, are not well-represented in the training data.  Such extreme events may be out of the observed distribution, so models that rely purely on correlations may struggle to produce accurate predictions in these rare (and potentially critical) situations.
Introducing an inductive bias that encodes knowledge of the underlying mechanisms generating the data can constrain inferences and lead to more robust predictions.

Mechanistic models alone, however, can oversimplify complex phenomena, making them overly sensitive to unmodeled nuisance variation.  Biomedical models, for instance, may summarize complex metabolic processes with relatively simple dynamics at a coarse level, ignoring sources of uncertainty due to noisy inputs or measurement error.
Such models are designed in highly controlled settings, and thus may be unable to cope with variation one might see outside the lab.

\begin{figure}[t]
\centering
    \begin{subfigure}{.5\textwidth}
        \centering
        \includegraphics[width=\textwidth]{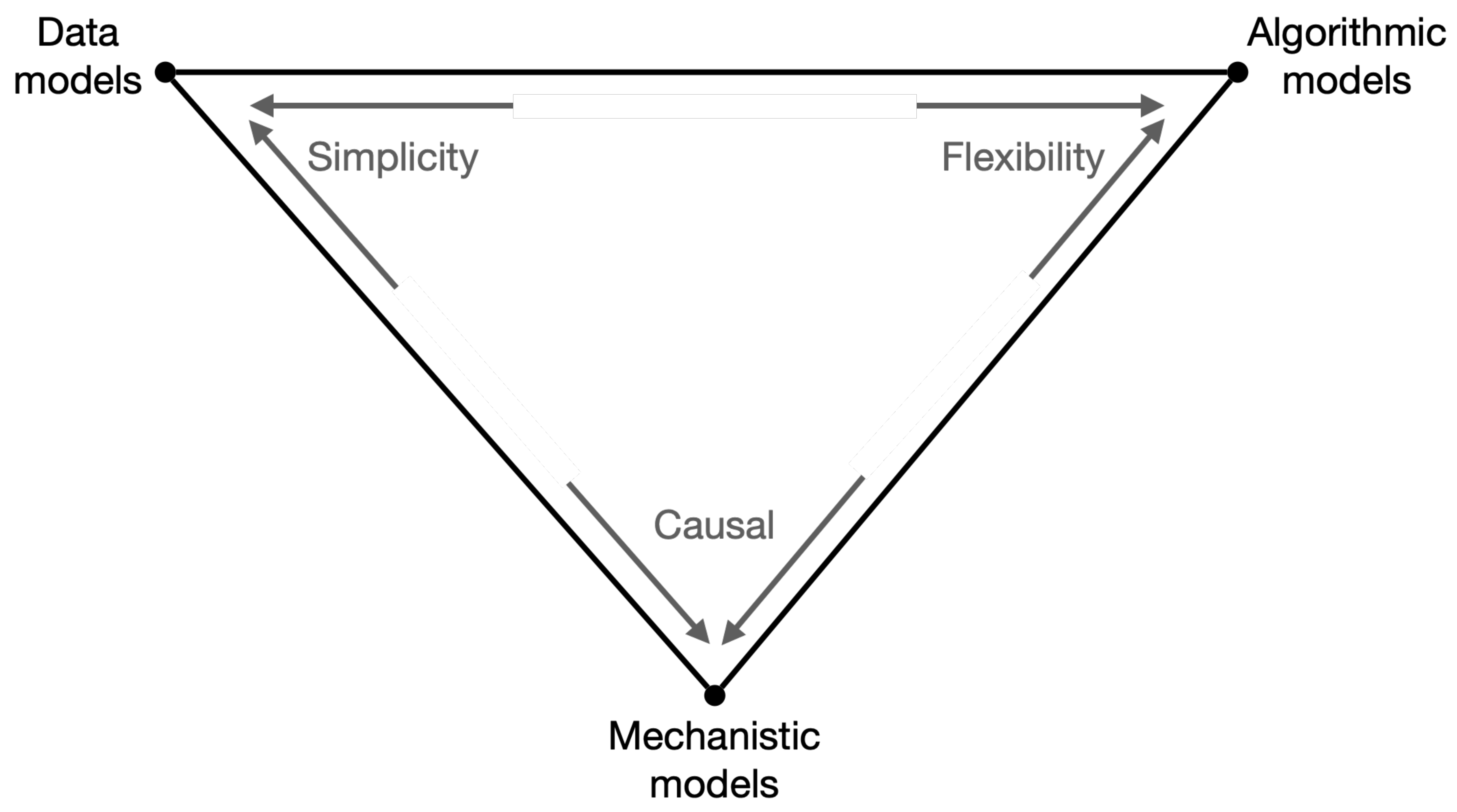}
        \caption{Conceptual model space.  Each extreme represents a preference --- simplicity, flexibility, or scientifically informed bias.}
        \label{fig:model-space}
    \end{subfigure}
    \quad~
    \begin{subfigure}{.42\textwidth}
        \centering
        \includegraphics[width=\textwidth]{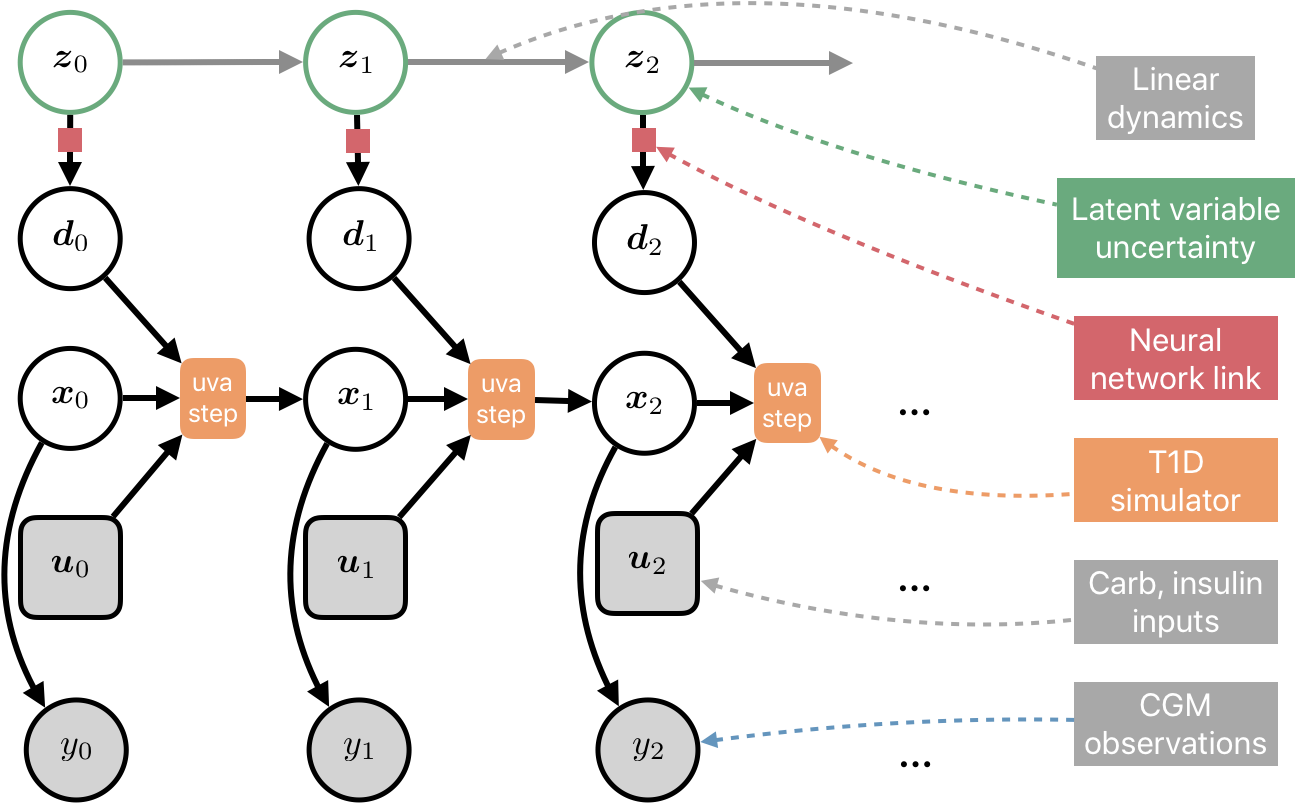}
        \caption{T1D hybrid model from \citet{miller2020learning}.  Linear dynamics and a flexible link function model time-varying simulator parameters.}
        \label{fig:t1d-hybrid}
    \end{subfigure}
\end{figure}

\paragraph{Exploring model space --- hybrid models.}
Data analysis requires striking a balance between model flexibility and an appropriate inductive bias, tailored to our goals.  Instead of limiting ourselves to one framework, we can explore the space of models between the extremes.  Movements in model space can introduce constraining assumptions, or conversely, flexibility.  Conceptually depicted in Figure~\ref{fig:model-space}, moving toward data models preferences simplicity; moving toward algorithmic models preferences flexibility; and moving toward mechanistic models preferences a scientifically-informed causal inductive bias.\footnote{Disclaimer: Mechanistic models aim to leverage a good approximation to the underlying causal mechanism that generates the data to form predictions. Formal causal inference requires more transparent assumptions to make causal statements, and can use data, algorithmic, and mechanistic modeling tools.}
The statistics and machine learning communities have been devising and experimenting with these hybrid models for decades now. 
And with modern computational tools, interpolation in the space of models between these three extremes has become more common. 

Modern machine learning models that blend the data and algorithmic paradigms include the structured variational autoencoder (sVAE) \citep{johnson2016composing}, and the output interpretable VAE (oi-VAE) \citep{ainsworth2018oi}. 
These models incorporate highly flexible neural network components together with a ``data model'' flavor --- inducing sparsity or using a structured probabilistic graphical model to describe a learned representation of the data.
Arguably, the LASSO \citep{tibshirani1996regression} lies somewhere between data and algorithmic models, as the procedure entails an algorithmic feature selection step, resulting in a simpler, easier-to-interpret linear model with (in many cases) mature statistical guarantees.

Another path for building more inductive bias into algorithmic models is to explore the algorithmic-mechanistic frontier.
One example is the physics-informed neural network, which introduces constraints on activations of the neural network to enforce predictions that are consistent with physical laws \cite{karpatne2017physics}.  
Equivariant neural networks are another approach that blend algorithmic and mechanistic models by encoding known symmetries in the data directly into the model \citep{cohen2016group}.  
Convolutional neural networks, arguably, also represent a step from algorithmic to mechanistic models by baking translation invariance into its architecture, compared to less constrained multi-layer perceptron models \citep{rumelhart1985learning}.
Even Breiman's description of pruning random tree nodes according to domain knowledge is a step from purely algorithmic models toward a more algorithmic-mechanistic model.

Applications in statistical neuroscience provide examples of models that blend the mechanistic and data paradigms.  
Time series models of neural image data can be constructed out of data model components (e.g., a linear dynamical system) carefully specified to approximate known biophysical models (e.g., Hodgkin-Huxley models of voltage propagation) to discover functional structure \citep{johnson2020probabilistic, sun2019scalable}. 

\paragraph{Glucose forecasts and T1D.}
A recent example from our own research focuses on glucose forecasts for type 1 diabetes (T1D), incorporating input information from a continuous glucose monitor (CGM), diet logs, and insulin delivered \citep{miller2020learning}.
Our goal was to form physiologically plausible and robust forecasts --- that is, sensible predictions in counterfactual contexts (e.g., under different insulin delivery or meal schedules).
We found that purely data and algorithmic models, like ARMA time series models and LSTM models, respectively, would often produce physiologically nonsensical counterfactual forecasts --- predicting large increases in glucose after a large insulin dose, for example. 

Alternatively, mechanistic models that simulate the insulin, glucose, and carbohydrate dynamics in the body have been developed to test insulin dosing strategies \citep{dalla2002oral, dalla2007meal, man2014uva}. 
These models have typically been used in simulation studies and developed in highly controlled settings.
Crucially, though, these models encode a sophisticated, scientifically-driven inductive bias for how insulin, glucose, and carbohydrates interact in T1D individuals.

However, we found that the T1D simulator alone is insufficient to describe real world CGM data.
The simulator, though encoding a physiology-based inductive bias, is too inflexible to handle sources of variation observable in real-world CGM data --- person-specific diurnal variation in insulin sensitivity, error in diet logs, and the individual's movement, among other reasons. 

We instead developed a hybrid model --- we allowed the simulator parameters to vary in time by introducing a latent linear dynamical system (data model) with a neural network link (algorithmic model) to map the latent variable to physiologically meaningful simulator parameters (mechanistic model). See Figure~\ref{fig:t1d-hybrid}.
Our approach obtained flexibility by allowing simulator parameters to vary in time, yet retained a strong inductive bias by essentially forcing predictions to use diet log and insulin inputs to explain CGM changes.  The simplicity of the linear dynamical system prior enabled fine control over the stability of the latent process.   

\paragraph{Modeling COVID-19 spread.}
Another example of hybrid modeling can be found in recent work focused on the COVID-19 pandemic.  
These models incorporate a new source of potentially relevant data, cell phone-based mobility time-series that describe movement of the population.
These hybrid models combine the mechanistic structure of the SEIR framework \citep{kermack1927contribution} with flexible components to incorporate time-varying mobility data to inform the transmission parameters, modulating the basic or effective reproductive number \citep{arik2020interpretable, flaxman2020estimating, liu2020estimating, miller2020learning}.
Some approaches incorporate time-varying indicators that describe certain public policy ordinances, such as school closings or stay-at-home orders, following a data model style \citep{flaxman2020estimating}.
The aim is to accurately predict future outbreaks and understand the effect of certain policies or population-level behavior changes on the spread of the disease.

This hybrid modeling framework is more flexible than the purely mechanistic modeling framework, resulting in better model fit.  
Even with the incorporation of mechanistic information, care must be taken when using these models to measure causal effects of interventions or produce long-term forecasts. 
In a thorough simulation study, \citet{soltesz2020effect} conclude that the model of \citet{flaxman2020estimating} is too flexible to reliably measure the impact of multiple non-pharmaceutical interventions in this observational setting, despite using a mechanistic epidemiological model component. 
Though the mechanistic model component constrains the results of the analysis to a smaller set of hypotheses than a purely algorithmic model, the hybrid model and data may still not pin down a satisfactory answer, requiring either further assumptions (which may or may not be appropriate) or the acquisition of better data. 

\paragraph{Conclusion.}
Developments in computational and algorithmic tools grant us finer control over model details than ever. 
Expanding Breiman's ontology to include mechanistic models --- and the space of hybrid models --- widens the set of questions we can ask and approaches we can take to answer them.
These models can incorporate a sophisticated scientific inductive bias (e.g., via a simulator as in the T1D case), but also retain the flexibility of algorithmic models to cope with unforeseen and unmodeled variation, and the simplicity of data models for statistical inference. 

For any particular application, a key question is, what is an appropriate operating point in hybrid model space?
To answer, we must compose the data, algorithmic, and mechanistic components to achieve a model that addresses the data analysis question at hand.
The optimal way to balance these components is an open question, but one that is specific to the goals of the application --- e.g., forming robust predictions of rare events.
Specific to the mechanistic component there are a number of challenges to address: How do we incorporate domain knowledge that is difficult to simulate, or with unknown structure?
Even given a causal model, can the data we observe identify the parameters necessary to form a new prediction or estimate an effect? 
How do we check or compare mechanistic model components?

Importantly, not all questions can be answered by any given dataset (particularly observational ones).  Sometimes, the data simply do not contain the information necessary to answer a particular query. Even if a mechanistic model gives you some traction --- i.e., a good set of structural assumptions to build upon --- misspecification and unidentifiability can lead you astray. 
Even within the hybrid modeling framework, bolstered by countless advances in computation, statistics, and machine learning, the simple insight from John Tukey remains timeless:~``The data may not contain the answer.'' \citep{tukey1986sunset}.

\bibliography{miller-foti-fox.bib}

\end{document}